\documentclass[a4paper,fleqn]{cas-sc}
\usepackage{apacite}
\usepackage{natbib}
\def\tsc#1{\csdef{#1}{\textsc{\lowercase{#1}}\xspace}}
\tsc{WGM}
\tsc{QE}
\tsc{EP}
\tsc{PMS}
\tsc{BEC}
\tsc{DE}
\begin{document}
\let\WriteBookmarks\relax
\def\floatpagepagefraction{1}
\def\textpagefraction{.001}

\shorttitle{Non-autoregressive Personalized Bundle Generation}

\shortauthors{Wenchuan Yang et~al.}

\title [mode = title]{Non-autoregressive Personalized Bundle Generation}                      
%\tnotemark[1,2]

%\tnotetext[1]{This document is the results of the research
%   project funded by the National Science Foundation.}

%\tnotetext[2]{The second title footnote which is a longer text matter
%   to fill through the whole text width and overflow into
%   another line in the footnotes area of the first page.}

\author[1]{Wenchuan Yang}[type=editor,
                        style=chinese,
                        auid=000,bioid=1,
                        orcid=0000-0003-3194-1690]

\fnmark[1]

% Email id of the first author\ead{wenchuanyang97@163.com}

%\ead[url]{www.cvr.cc, cvr@sayahna.org}

\credit{Conceptualization of this study, Methodology, Software, Writing - Original Draft}

\affiliation[1]{organization={College of Systems Engineering, National University of Defense Technology},
    %addressline={Radarweg 29}, 
    city={Changsha},
    % citysep={}, % Uncomment if no comma needed between city and postcode
    postcode={410073}, 
    country={PR China}}

\author[2]{Cheng Yang}[style=chinese, orcid=0000-0001-7821-0030]

\credit{Writing - Review & Editing, Methodology}

\affiliation[2]{organization={School of Computer Science, Beijing University of Posts and Telecommunications},
    %addressline={Radarweg 29}, 
    city={Beijing},
    % citysep={}, % Uncomment if no comma needed between city and postcode
    postcode={100080}, 
    country={PR China}}

\author[1]{Jichao Li}[style=chinese, orcid=0000-0002-1781-3732]

\credit{Data Curation}

\author[1]{Yuejin Tan}[style=chinese]

\credit{Validation, Funding acquisition}

\author[1]{Xin Lu}[style=chinese, orcid=0000-0002-3547-6493]

\cormark[1]
\ead{xin.lu.lab@outlook.com}
\credit{Supervision, Resources}

\author[2]{Chuan Shi}[style=chinese, orcid=0000-0002-3734-0266]
\cormark[1]
\ead{shichuan@bupt.edu.cn}
\credit{Supervision, Writing - Review & Editing}

\cortext[cor1]{Corresponding author}
%\cortext[cor2]{Principal corresponding author}

%\fntext[fn1]{This is the first author footnote. but is common to third
 % author as well.}
%\fntext[fn2]{Another author footnote, this is a very long footnote and
 % it should be a really long footnote. But this footnote is not yet
  %sufficiently long enough to make two lines of footnote text.}

%\nonumnote{This note has no numbers. In this work, we demonstrate $a_b$
%  the formation Y\_1 of a new type of polariton on the interface
%  between a cuprous oxide slab and a polystyrene micro-sphere placed
%  on the slab.
%  }

\begin{abstract}
The personalized bundle generation problem, which aims to create a preferred bundle for user from numerous candidate items, receives increasing attention in recommendation. However, existing works ignore the order-invariant nature of the bundle and adopt sequential modeling methods as the solution, which might introduce inductive bias and cause a large latency in prediction. To address this problem, we propose to perform the bundle generation via non-autoregressive mechanism and design a novel encoder-decoder framework named BundleNAT, which can effectively output the targeted bundle in one-shot without relying on any inherent order. In detail, instead of learning sequential dependency, we propose to adopt pre-training techniques and graph neural network to fully embed user-based preference and item-based compatibility information, and use a self-attention based encoder to further extract global dependency pattern. We then design a permutation-equivariant decoding architecture that is able to directly output the desired bundle in a one-shot manner. Experiments on three real-world datasets from Youshu and Netease show the proposed BundleNAT significantly outperforms the current state-of-the-art methods in average by up to 35.92\%, 10.97\% and 23.67\% absolute improvements in Precision, Precision+, and Recall, respectively.
\end{abstract}

% \includegraphics{figs/grabs.pdf}

%\item Research highlights item 1
%\item Research highlights item 2
%\item Research highlights item 3

% Keywords

\maketitle

\section{Introduction}
The recommender system has now proved an effective tool for alleviating the information overload phenomenon in our daily life \citep{duan2023multi,sheng2023enhanced, hu2020graph}. Generally, a recommender system predicts whether the target user will be interested in the item or not, and then returns a list of items from a vast candidate pool for potential choices. By doing so, users’ demand could be satisfied to the maximum extent. Other than recommending the single item to users, delivering a size-K ($K\ge2$) item-set named bundle becomes a common practice in online services  \citep{zhu2014bundle,li2020survey,kouki2019product}. The bundle has two key characteristics: 1) items should be appealing to target users, and 2) items should be compatible with each other and express the same topic or style. Some typical forms of the bundle are the playlist on Netease music platform, the game collection on Steam platform, and product set on shopping website Taobao. 

The advantages of delivering a bundle are easily recognized \citep{pathak2017generating,chen2019matching,ma2022crosscbr,zhang2022suger,yang2023heterogeneous,ding2023personalized}. For users, item bundling could better tailor to the needs by helping users find surprisingly interested items. For service providers, recommending bundles could expose more items to users. Particularly, in online e-commerce, with an attractive discount rate, the bundling strategy could even possibly increase sales \citep{liu2017modeling,sun2021product}.

There exist two lines of research on recommending bundles, one called pre-built bundle recommendation is in line with top-K recommendation which aims to rank the most likely preferred bundle (already existed) for users. The other called personalized bundle generation task is to investigate how to select items from the candidate set to composite suitable bundles for users as shown in Figure 1a, which is the focus of our paper.

Although a decent portion of the literature focuses on the pre-built bundle recommendation \citep{ma2022crosscbr,zhang2022suger,chen2019matching,chang2021bundle,vijaikumar2021gram}, relatively little effort has been made to push forward the development of bundle generation techniques, leaving a large space waiting for exploration. Existing works \citep{gong2019exact,deng2021build} generally model the bundle as a sequence and perform the generating process in an autoregressive manner, in which the items are selected one by one. And they propose to utilize sequential methods like PointerNet \citep{vinyals2015pointer} or reinforcement learning techniques \citep{schulman2017proximal} to resolve it. Despite their effectiveness, we suggest that sequential modeling is still insufficient to find the optimal bundle. The reason is that it ignores the order-invariant property embedded in the bundle. 

 \begin{figure*}[!t]
  \setlength{\belowcaptionskip}{0em}
  \centering
  \includegraphics[width=\linewidth]{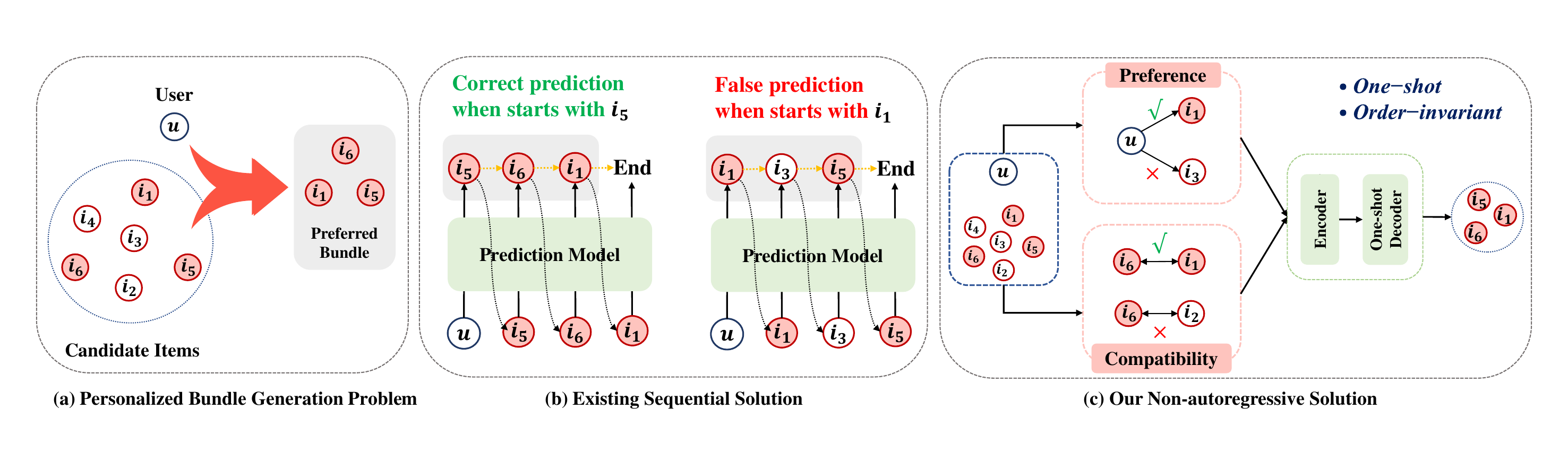}
  \caption{A toy example illustrating personalized bundle generation. Figure 1a: Given user $u$ and candidate items $i_{1},...,i_{6}$, the goal is to 
 generate the preferred bundle consists of $i_{1}$, $i_{5}$, and $i_{6}$. Figure 1b shows the 4-step generating process based on sequence modeling, which is unaware of multiple optimal sequential orders and might result in inference failure when changing the sequential order. Figure 1c illustrates the proposed non-autoregressive generation which aims to output the size-3 bundle in 1 step by utilizing preference and compatibility information (An encoder-decoder architecture is adopted in our paper).}
\end{figure*}

To be more specific, we detail the generation process based on sequence modeling in Figure 1b. To search for the targeted $\left\{i_{1}, i_{5}, i_{6}\right\}$ bundle, it follows a 
 step-by-step manner in which $i_{6}$ follows $i_{5}$, $i_{1}$ selected right after $i_{6}$, and $\left\{i_{5}-i_{6}-i_{1}\right\}$ is assumed to be the best ordering to recovering the bundle. However, there are $3!$ equally good solutions for the size-$3$ bundle \citep{zaheer2017deep,zhang2019deep,sui2023joint},
 when we place $i_{1}$ in the first place, the model could be confused since no dependency information is available, as $i_{1}$ is the last predicted item before. Hence, relying on a certain ordering might introduce unnecessary inductive bias and harm the generalization ability of the model. Besides, it requires 4 times inference for a size-$3$ bundle indicating the large latency in sequential prediction.

Aware of these setbacks, in this study, the bundle is structured into a set and we propose to perform the bundle generation via non-autoregressive mechanism which attempts to output the desired items in one-shot. Though the non-autoregressive solution seems like a straightforward approach as depicted in Figure 1c, the problem remains highly non-trivial with the following unique challenges:
\begin{itemize}
    \item How to make the generation process aware of the compatibility? Different from the sequential modeling in which compatibility could be promised by conditioning on the preceding ones, the non-autoregressive generation process is a one-shot process, consequently, we need to find a way to ensure the item compatibility globally.
    \item How to design a proper decoding approach that is equivariant to the permutations of constituent items? The main difficulty of non-autoregressive decoding lies in the property that items within the bundle are freely interchangeable. Imposing an inherent order \citep{gong2019exact,deng2021build} made it easier, but also can make the prediction highly sensitive to the input order.
\end{itemize}

To address these challenges, inspired by non-autoregressive (NAT) style transformers, we here propose a novel encoder-decoder framework \footnote{Although the proposed framework is mainly based on existing popular techniques, the novelty of our work lies in the effective combination along with scenario-specific modifications tailored for non-autoregressive generation. The empirical results further demonstrate the effectiveness and efficiency of proposed BundleNAT} for bundle composition named BundleNAT. Our proposed solution efficiently facilitates both user-based preference signal and item-compatibility signal as the overall dependency pattern and pairs it 
with a bundle-specific non-autoregressive decoding network. In detail, we first identify two key factors, i.e., user-based preference and compatibility among items, crucial for personalized generation. Inspired by pre-training techniques \citep{devlin2018bert,zhang2019ernie,shen2023towards,nowakowski2023adapting}, we decide to utilize a conventional recommendation model to learn preference signals based on user-item interactions. For compatibility signals, since there is no ground-truth data on the relation of substitution and complement among items, we propose employing the co-occurrence relation as the approximation and use the graph neural network (GNN) as the extractor to capture compatibility signals. By employing a self-attention based encoding network, the model is able to further learn the global dependency patterns essential for decoding. Secondly, inspired by NAT transformer studies \citep{gu2017non,huang2022learning}, we propose to adopt a non-autoregressive decoding mechanism, which is independent of specific ordering, to recover the bundle. However, simply applying the vanilla non-autoregressive decoding mechanism to the bundle generation scenario is not feasible. The reason is that the \emph{multi-modality} issue \citep{zhan2022non,jiang2021improving,niwa2023nearest,ma2023fuzzy}, which refers to the inability to identify multiple equal combinations of items owing to the parallel output, is not solved and will 
 result in poor generation in our scenario. To tackle the problem, we then propose a permutation-equivariant decoding network to alleviate this issue, meanwhile improving the performance by retrieving a base from the encoder via a novel copy mechanism. Lastly, considering the order-invariant feature of bundle, order-agnostic cross entropy \citep{du2021order} is applied to guide the model training. Extensive experiments conducted on three real-world datasets validate the superiority of the proposed BundleNAT against state-of-the-art methods and the significance of the designed modules. The salient points of this work are:

\begin{itemize}
    \item We take the first step to formulate the bundle generation task via the non-autoregressive mechanism, which is a more proper way to facilitate the order-invariance property of the bundle.
    \item We propose a novel non-autoregressive mechanism featured framework for personalized bundle generation, which can capture both the preference and compatibility pattern. We further design a bundle-specific decoding process combined with a copy mechanism to recover the targeted bundle.
    \item The experimental results demonstrate the effectiveness and efficiency of our proposed framework over existing state-of-the-art, achieving average 35.92\%, 10.97\%, and 23.67\% absolute performance gain on Precision, Precision+, and Recall against the second-best baseline, respectively. 
\end{itemize}

\section{Related Work}
In this section, we briefly review the existing relevant literature in the personalized bundle generation domain.
\subsection{Bundle Generation}
The core idea of bundle generation is to develop a model that is capable of picking up a desired set of compatible items from a vast candidate pool. Early works mainly adopt utility function and statistical methods to infer the target bundle. Xie et al. \citep{xie2014generating} proposed a linear additive utility function based on implicit user feedback to generate personalized bundles, however, the function is unable to model complex dependencies among items. Ge et al. \citep{ge2017effects} have studied the bundling strategy on e-commerce platforms and found that items with more reviews, particularly those with photos attached, are more likely to be included in the bundle. Liu et al. \citep{liu2017modeling} designed a probabilistic model named BPM to learn composition factors based on users' buying motives, and the preferred bundle is generated via finding complementary items concerning the target item.

Later, Pathak et al. \citep{pathak2017generating} built a greedy generation strategy upon the Bayesian Pairwise Ranking (BPR) \citep{rendle2012bpr} framework. The strategy first trained a bundle preference model, then the best bundle is picked from the candidate pool containing random variants of the initial bundle based on preference score. Vijaikumar et al. \citep{vijaikumar2021gram} developed a monotone submodular function for scoring generated bundles and applied a greedy algorithm to approximate the optimality. However, both the generation process follows a pure heuristic way, which might result in unaffordable solving time cost in real-world practice.

Recently, Gong et al. \citep{gong2019exact} interpreted the generation task as a maximal clique optimization problem on an item-item graph. It factorized the problem in a sequential modeling manner and then combined the reinforcement learning (RL) mechanism with an encoder-decoder framework to perform generation. In line with \citep{gong2019exact}, Deng et al. \citep{deng2021build} presented the problem as a multi-step Markov Decision Process and proposed a pure RL framework. The method first constructed a user-item preference and item compatibility model, then facilitated corresponding feedback signals along with the evaluation metrics as the reward function following curriculum training methodology. In a closely related work, Wei et al. \citep{wei2022towards} examined the superiority of non-autoregressive decoding manner and applied it to the bundle creative generation scenario, which takes heterogeneous objects into account. The method utilized a standard non-autoregressive encoder-decoder architecture in the NLP field, and considered an extra contrastive learning objective to further ensure generation quality.

However, our work is different from theirs in several main points:

\textbf{Firstly}, we are fully aware of the order-invariant nature of bundles and propose a novel non-autoregressive transformer architecture as the solution. While \citep{gong2019exact,deng2021build,bai2019personalized} identifying the bundle as a sequence and utilizing sequential modeling methods as the solution, the generation performance will be affected by the contradiction between the order-invariant nature of the bundle and presumed optimal sequential order. Besides, the sequential methods generally suffer from long-term bottleneck and inference latency.

\textbf{Secondly}, we propose a proper way to encode the intrinsic compatibility among items into generating a feasible bundle, which is either ignored or insufficiently encoded in existing works. For example, in \citep{deng2021build}, each bundle is seen as a sentence, and word2vec \citep{mikolov2013efficient} is applied to obtain item compatibility, however, the bundle has no strict order like sentences and context-based learning only captures local dependency while items within the bundle are globally correlated.

\textbf{Thirdly}, as a similar task, bundle creative generation \citep{wei2022towards} aims to find a set of heterogeneous items that satisfy users’ preference to the utmost. To improve the generation efficiency, the vanilla non-autoregressive architecture is directly adopted. In this study, the motivation comes from the order-invariant nature of the bundle rather than seeking superior time efficiency. Meanwhile, the non-autoregressive decoding method used in \citep{wei2022towards} is not applicable in bundle generation due to the ignorance of item compatibility and the multi-modality issue.

\textbf{Lastly}, we experiment with our model on real-world bundle recommendation datasets which truly demonstrate the characteristics of the bundle. In previous works, ground-truth bundle data is normally sampled from users’ historical purchases in which totally irrelevant items have a great probability curated as a whole, generally containing lots of noisy information \citep{tzaban2020product}. The experimental results are more reliable to show the effectiveness of the architecture design.

\subsection{Bundle Completion}
There is another line of work focusing on the personalized bundle generation task, in which the generation is essentially interpreted as a completion process. Thus, the goal of these studies switched to finding the next suitable item based on the incomplete existing bundle. For example, Bai et al. \citep{bai2019personalized} worked on a bundle-list recommendation problem and proposed a bundle generation network based on Long Short-Term Memory (LSTM) \citep{hochreiter1997long} and Determinantal Point Process \citep{chen2018fast} selection to find high-quality and diversified bundles. Chang et al. \citep{chang2021bundle} proposed to treat the bundle as a sparse-connected item graph and designed a GNN-based model named BGGN to predict both the right nodes and edges in the graph. Jeon et al. \citep{jeon2023accurate} simplified bundle generation as the recovering process from the incomplete one and designed a neural network to learn the user-bundle (incomplete) pair embedding so that the rest of the items could be correctly retrieved.

Methods for bundle completion are excluded from further discussion in this paper, due to the totally different problem settings and incompatible hypothesis.

\section{Problem Formulation}
Given a candidate items set $I=\left\{v_1, v_2, \ldots, v_n\right\}$ with \emph{n} stands for the set size, the goal of personalized bundle generation is to find the specific item-set $B_u=\left\{v_j \mid j=1,2, \ldots, K, v_j \in I\right\}$, so that $B_u$ is most likely the correct and preferred combination of items for each user \emph{u}. We therefore derive the problem formulation as follows,
\begin{equation}
\max \sum_u P\left(B_u \mid u, I\right)
\end{equation}
where $P\left(B_u \mid u, I\right)$ measures the probability of generated $B_u$ being satisfied. However, simply conditioning on user preference cannot ensure the rationality of the generated bundle, we introduce item compatibility information to further supervise the generation process:
\begin{equation}
\max \sum_u P\left(B_u \mid u, I, O\right)
\end{equation}
where $O$ represents the compatibility pattern. 

Inspired by the order invariant property of the bundle, we can therefore model the task in non-autoregressive manner, in which the sequential order is no longer needed for generating. We should be able to find the optimal bundle by predicting an item set from the candidate pool in one-shot.

\begin{figure*}[!h]
  \centering
  \includegraphics[width=\linewidth]{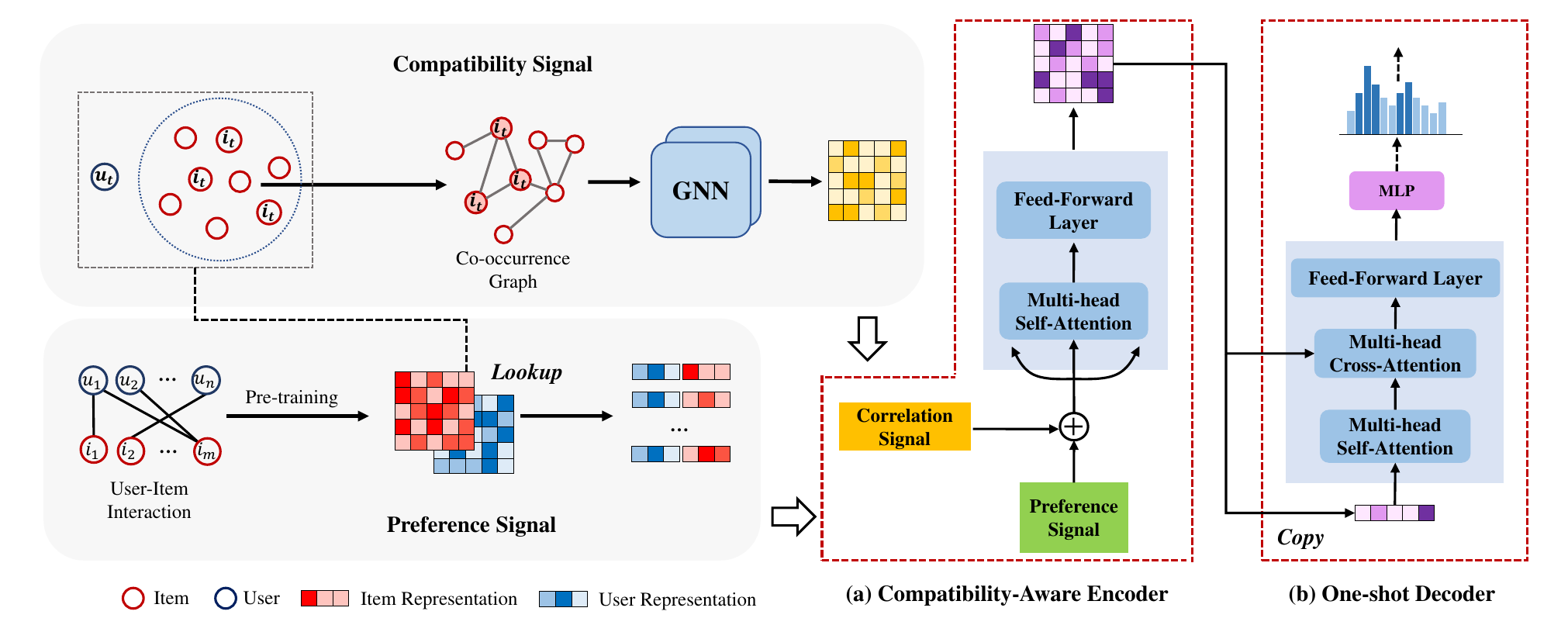}
  \caption{The overall architecture of the proposed framework. $u_{t}$ stands for the target user, the items within the ground-truth bundle are marked by $i_{t}$.}
\end{figure*}

Thus, the probability of inducing a bundle could be factorized as follows.
\begin{equation}
P\left(B_u \mid u, I, O\right)=\sum_{j=1, v_j \in I}^K p\left(v_j \mid u, O\right)
\end{equation}

Relying on this factorization, the entire generation process is fully aware of users’ preference and items’ compatibility. Meanwhile, the formulation is order-invariant since the summing operation is commutative, the prediction remains constant no matter how the order changes.

\section{Proposed Framework}
In this section, more details are presented regarding the proposed BundleNAT framework. We first give an overview of the framework, and then we further clarify the design of two main modules of NAT, i.e., compatibility-aware encoder and one-shot decoder.

\subsection{Overview}
The overall architecture of the proposed framework is shown in Figure 2. The BundleNAT falls into two main parts: compatibility-aware encoder and one-shot decoder. For the encoder, we have introduced two kinds of signal to learn the needed global dependency information, i.e., the preference signal and the item compatibility signal. We propose to capture item compatibility signal from the co-occurrence graph to strengthen the intra-relatedness within the bundle, meanwhile, the preference signal extracted from user-item interactions is utilized to ensure each item retrieved is appealing to the user. Then a self-attention based encoding network is employed to further learn the global dependency pattern. When designing the decoding strategy, inspired by vanilla NAT framework, we revise the original decoder architecture from parallel decoding into a one-shot decoding manner. Besides, a copy mechanism is proposed to guide the decoding process towards the optimal solution and an MLP-based projection module is added to enable the model to directly output the item set from the predicted distribution. In the end, we utilize order-agnostic cross entropy as the training loss to guide model optimization.

\subsection{Compatibility-Aware Encoder}
\emph{Preference signal}. The generated bundle should well satisfy users' need, which means each item within the bundle is favorable. Given the candidate item-set $I$ and user $u$, we facilitate the preference signal \textbf{\emph{P}} to ensure the personalized property, which is derived as follows:
\begin{equation}
\begin{aligned}
& \boldsymbol{P}=\left[p_1, p_2, \ldots, p_n\right] \\
& \text { with } p_i=e_{v_i} \oplus e_u
\end{aligned}
\end{equation}
where $p_i \in \boldsymbol{R}^d$ represents the preference signal embedded in $u-v_i$ pair, $e_{v_i}\in I$, $e_u$ denote feature vectors for item $v_i$ and user $u$ respectively, $\oplus$ refers to concatenation operation. Here, $e_{v_i}$, $e_u$ are learned by matrix factorization model optimized by BPR \citep{rendle2012bpr} based on user-item interaction data. Consequently, $p_i$ can migrate the user’s preference into item selection for the bundle. Notably, the preference model could be replaced by other similar models like Wide-and-Deep \citep{cheng2016wide} and LightGCN \citep{he2020lightgcn}.

\emph{Compatibility signal}. The key challenge in non-autoregressive bundle generation is how to guarantee the rationality of the selected item-set, in other words, constitute items should be compatible. Selection only relies on preference could possibly result in meaningless bundle in which incompatible items could appear together. Here, we first construct a co-occurrence graph $G$ that implies the item compatibility pattern. Specifically, for each item $v_i$ in the candidate set $I$, there is an occurrence frequency vector $f_{v_i} \in {R}^{N_b}$ retrieved from ground-truth bundle-item affiliations where $N_b$ denotes the number of bundles. Specifically, $G$ is built as follows,
\begin{equation}
\begin{gathered}
G=\boldsymbol{D}^{-1 / 2} \cdot\left(\boldsymbol{F} \cdot \boldsymbol{F}^T\right) \cdot \boldsymbol{D}^{-1 / 2} \\
\boldsymbol{F}=\left[f_{v_1}, f_{v_2}, \ldots, f_{v_n}\right]^T, G \in \boldsymbol{R}^{n \times n}
\end{gathered}
\end{equation}
where \textbf{\emph{D}} is the degree matrix of $\boldsymbol{F} \cdot \boldsymbol{F}^T$ utilized for normalization, $g_{ij} \in G$ shows the compatibility degree between item $v_i$ and item $v_j$. Thus, we adopt the GNN model to transform the degree information into vectorized representation $c_{v_i} \in {R}^d$,
\begin{equation}
c_{v_i}=\sum_{v \in N\left(v_i\right)} G N N_G\left(v_i\right)
\end{equation}
where $N(v_i)$ stands for the neighbor set of item $v_i$ including itself. The detailed learning mechanism is expressed as,
\begin{equation}
\begin{gathered}
c_{v_i}^{(k)}=\sigma\left(W_{(k)}\left(c_{v_i}^{(k-1)}+\operatorname{Agg}\left(c_v^{(k-1)}\right)\right)+b_{(k)}\right) \\
c_{v_i}{ }^{(0)}=z_{v_i}
\end{gathered}
\end{equation}
where $k$ means the $k$th propagation layer, $W_{(k)}$, and $b_{(k)}$ are the corresponding learnable weight matrix and bias, respectively. $c_{v_i}^{(k-1)}$ stands for the preceding learned representation of $v_i$, $c_{v}^{(k)}$ is the representation of $v_i$’s neighboring nodes, $Agg$ stands for the weighted aggregation operation, with the weight decided by $g_{ij}$. $z_{v_i}$ denotes the initial trainable node feature of $v_i$.

\emph{The input}. Positional encoding is one of the most successful designs in Transformer architecture that helps the encoder network aware of the position information of tokens, thus enhancing the performance in various language tasks \citep{vaswani2017attention,liu2019roberta,radford2019language}. Here, we employ the Absolute-Position-Encoding-style (APE-style) input as the backbone, in which the compatibility signal is seen as the position encoding for the candidate items. Although simple, APE enjoys a great property proved by \citep{luo2022your,yun2019transformers} that Transformer architecture with APE is capable of approximating any continuous sequence-to-sequence function in a compact domain. Meanwhile, Relative Position Encoding fails in this theorem because the positional information will be suppressed into stochastic signals when placed in softmax exponentiation. 
Thus, in the proposed BundleNAT framework, we derive the input for encoding as $\boldsymbol{X} \in \boldsymbol{R}^{n \times d} = \boldsymbol{P} + \boldsymbol{C}$, where 
$\mathbf{C} = \left\lbrack {c_{v_{1}},c_{v_{2}},\ldots,c_{v_{n}}} \right\rbrack^{T}$.

\emph{Encoding network}. We adopt an attention-based encoding network to learn a solid dependency pattern for subsequent decoding. The basic unit of the encoding network consists of two kinds of layers: a self-attention layer $Attention$ followed by a point-wise feed-forward layer $FFN$. Given the input $\boldsymbol{X}$, the workflow of a unit is defined as follow,
\begin{equation}
\begin{gathered}
\boldsymbol{X_A}=Attention(\boldsymbol{X})=s o f t m a x\big(\frac{\big(W_{Q}\boldsymbol{X}\big)\cdot(W_{K}\boldsymbol{X}\big)^{T}}{\sqrt{d}}\big)\cdot(W_{V}\boldsymbol{X}) \\
\boldsymbol{X_F}=FFN(\boldsymbol{X})=\boldsymbol{X_A}+\sigma(\boldsymbol{X_A}W_{1})W_{2} 
\end{gathered}
\end{equation}
where $W_{Q}$, $W_{K}$ and $W_{V}$ in self-attention layer are learnable projection vectors, $d$ stands for the embedding dimension of $\boldsymbol{X}$. $W_{1}$ and $W_{2}$ are learnable weight matrices in feed-forward layer, with activation function $\sigma$ set to be $relu$ function. Here, multi-head attention is further adopted to capture different dependency patterns among the candidate set, those learned patterns will then be concatenated as the output,
\begin{equation}
    \begin{gathered}
 \boldsymbol{X_A}=Multihead~Atention(\boldsymbol{X})=(head_1\oplus head_2\oplus...\oplus head_k)\cdot W_M \\
 head_{i}=softma x\big(\frac{\big(W_{Q}^{i}\boldsymbol{X}\big)\cdot\big(W_{K}^{i}\boldsymbol{X}\big)^{T}}{\sqrt{d}}\big)\cdot(W_{V}^{i}\boldsymbol{X}\big) 
\end{gathered}
\end{equation}
where $W_{Q}^{i}$, $W_{K}^{i}$ and $W_{V}^{i}$ are learnable projection vectors for $i$-th head, $W_M$ denotes the parameterized transformation matrix, $\oplus$ stands for concatenation.

\subsection{One-Shot Decoder}
Non-autoregressive decoding has received lots of attention in the NLP field in recent years \citep{ren2020study,xiao2023survey}, different from auto-regressive decoding (or sequential decoding), non-autoregressive decoding aims to output tokens parallel so that the inference latency is largely reduced. In this study, we suggest a sequential order is not necessary for the generation by recognizing the order-invariant nature of the bundle. Therefore non-autoregressive decoding mechanism becomes a natural option for bundle composition. However, directly utilizing the vanilla non-autoregressive decoding from NLP is not feasible resulting from two challenges: (1) multi-modality, referring to the phenomenon that parallel output is unaware of the multi-combination distribution of target bundle. For example, $\left\{1,3,5,7\right\}$ along with $\left\{5,3,1,7\right\}$ represents the same bundle, for parallel decoding it might lead to $\left\{5,3,5,7\right\}$ due to independence of multi-channel prediction; (2) the performance degradation attributes to the removal of sequential dependency in prediction. 

To deal with the challenging issue, we here revise the vanilla NAT architecture and propose a one-shot decoder with a copy mechanism to directly output the desired item set.

\emph{The input}. As demonstrated by \citep{niwa2023nearest}, the poor performance of NAT models could resulted from the from-scratch decoding process, which means if the decoder starts from low-quality input, it’s difficult to reach the demanded output even through several iterations. Besides utilizing the learned global dependency pattern to guide the decoding, we suggest a high-quality “start” is also vital for further improving the performance and generating the satisfied bundle effectively. Subsequently, we propose to copy the output $\boldsymbol{X_F}\in {R}^{n \times d}$ from the encoder via a mean pooling function, to get abstract global dependency feature, which serves as the start point for decoding,
\begin{equation}
    h_m=\frac{1}{|P_m|}\sum_{(i,j)\in P_m}x_{ij},m=1\text{,}2,...,d
\end{equation}
where $x_{ij}\in \boldsymbol{X_F}$, $P_m$ denotes the pooling area to learn $h_m$ which belongs to $\boldsymbol{R}^{n \times 1}$, and we have $d \cdot P_m \in {R}^{n \times d}$. $|\cdot|$ indicates the number of elements in the area, and the input for the decoder is formulated as $\boldsymbol{h}=\left\{h_m \mid m=1,2,…,d\right\},\boldsymbol{h} \in {R}^{1 \times d}$. There is a twofold advantage by reducing the input to ${R}^{1 \times d}$: firstly, the model no longer needs to match the predictions with the items \citep{zhang2019deep}, therefore avoiding multi-modality issue; secondly, the decoder works for variable-size bundles, while the vanilla method needs to change the input as the size varies.

\emph{Decoding network}. The basic block of proposed decoder contains three kinds of layers: one-token self-attention layer, cross-attention layer, and feed-forward layer. Specifically, the one-token self-attention layer is derived as follows:
\begin{equation}
    \boldsymbol{h}^{\prime}=A t t e n t i o n(\boldsymbol{h})=s o f t m a x\big(\frac{\big(W_{Q}^{d1}\boldsymbol{h}\big)\cdot\big(W_{K}^{d1}\boldsymbol{h}\big)^{T}}{\sqrt{d}}\big)\cdot\big(W_{V}^{d1}\boldsymbol{h}\big)
\end{equation}
where $W_{*}^{d1} \in {R}^{d \times d}$ is the parameterized transformation vector. The cross-attention layer is defined based on the interaction between encoder output and decoder input, the intuition is to suppress the learned fine-grained dependency information into decoding process,
\begin{equation} 
\boldsymbol{h}''=CrossAttention(\boldsymbol{h'})=softmax\bigl(\frac{\left(W_Q^{d2}\boldsymbol{X_F}\right)\cdot\left(W_K^{d2}\boldsymbol{h}'\right)^T}{\sqrt{d}}\bigr)\cdot\left(W_V^{d2}\boldsymbol{h}'\right)
\end{equation}
where $W_{*}^{d2} \in {R}^{d \times d}$ is the trainable parameter matrix, and the learned  compatibility $\boldsymbol{X_F}$ is served as the \emph{query} vector. $\boldsymbol{h}'' \in {R}^{1 \times d}$ is then fed into a feed-forward layer, and obtains the output representation $\boldsymbol{h_d}$:
\begin{equation}
\boldsymbol{h_d}=\boldsymbol{h}''+\sigma(\boldsymbol{h}''W_{1}^{d3})W_{2}^{d3}
\end{equation}
where $W_{*}^{d3} \in {R}^{d \times d}$ denotes the parameterized weight matrix. 

Notably, multi-head attention mechanism is also utilized in both one-token self-attention layer and cross-attention layer.

\emph{Prediction}. We further add an MLP-based network module to project back to the full-size item-set and return the predicted distribution, consequently, we can pick the desired item directly instead of inferring the item position \citep{wei2022towards}.
\begin{equation}
    \boldsymbol{h}_{o}=\sigma(\boldsymbol{h}_{d}W_{o}+b_{o})
\end{equation}
where $W_o$ is a learnable parameter matrix whose dimension is defined as $n \times N$, $N$ denotes the number of all items, $b_o$ is the bias factor, $\boldsymbol{h}_o \in \boldsymbol{R}^{1 \times N}$ is the predicted distribution for final inference.
\subsection{Model Optimization}
For inference, we use \emph{argmax} to produce the size-K bundle based on $\boldsymbol{h}_{o}=\{h_{o}^{1},h_{o}^{2},...,h_{o}^{N}\}$. 
For loss calculation, cross-entropy loss is a natural choice for training the structured prediction problem, however, in bundle generation, the strictly aligned cross entropy would falsely penalize the inference due to the ignorance of order-invariant property. Inspired by \citep{du2021order}, we propose a bundle-specific prediction learning objective to guide the model training. To illustrate, the inferred items are denoted as $\tilde{B}=\{v_{b_i}|b_i\in N,|b_i|=k\}$, the corresponding predicted distribution is denoted by $\tilde{b}=\{h_o^{b_i}\}$. We have an ordering space $\boldsymbol{B}=\{\tilde{B}_{(1)},\tilde{B}_{(2)},\ldots,\tilde{B}_{(N!)}\}$ which contains $N!$ kinds of permutation of predicted items, the bundle generation objective is formulated as retrieving the best ordering $\tilde{B}_{(i)}$ to minimize the cross-entropy loss,
\begin{align}
    L_{bundle} &=arg\operatorname*{min}_{\tilde{B}(i)\in B}\left(XE\left(\tilde{B}_{(i)},Y\right)\right) \\
    &=arg\operatorname*{min}_{\tilde{B}(i)\in B}\left(-\sum_{y_i\in Y}y_i log\left(h_o^{b_i}\right)+(1-y_i)log\left(1-h_o^{b_i}\right)\right)
\end{align}
where $XE(\cdot)$ is the standard cross entropy loss, $Y$ is the ground-truth bundle, $y_i$ represents each label belongs to $Y$. Specifically, to search for the best ordering $\tilde{B}_{(i)}$ efficiently, we leverage the Hungarian Matching algorithm \citep{kuhn1955hungarian}.

\section{Experiments}
In this section, we detail the experimental settings and present empirical results to demonstrate the effectiveness of the proposed BundleNAT. The experiments answer the following Research Questions (RQs):
\begin{list}{\labelitemi}{\leftmargin=1em}
    \setlength{\topmargin}{0pt}
    \setlength{\itemsep}{0em}
    \setlength{\parskip}{0pt}
    \setlength{\parsep}{0pt}
    \item \textbf{RQ1}: Does the proposed BundleNAT yield better generation performance compared with existing state-of-the-art?
    \item \textbf{RQ2}: How do different components (\emph{e.g.}, compatibility signal extraction) contribute to the performance of BundleNAT?
    \item \textbf{RQ3}: How well is the efficiency of BundleNAT for bundle generation?
    \item \textbf{RQ4}: How do different hyper-parameters (\emph{e.g.}, the depth of encoding/decoding network) affect the performance of BundleNAT?
\end{list}

\subsection{Datasets}
The evaluation datasets for the experiment are constructed based on two popular real-world bundle recommendation datasets, and the detailed statistics of original and constructed datasets are stated in Table 1.
\begin{list}{\labelitemi}{\leftmargin=1em}
    \setlength{\topmargin}{0pt}
    \setlength{\itemsep}{0em}
    \setlength{\parskip}{0pt}
    \setlength{\parsep}{0pt}
    \item \textbf{Youshu}. The dataset is collected by \citep{chen2019matching} from Youshu, a book-review website in China. Every bundle is a list of books that users might be interested in. To obtain the standard dataset for evaluation, we first create the ground-truth recommended size-$K$ bundles by randomly sampling $K$ items from each interacted bundle for a user, ensuring the size-$K$ bundles a user interacted with are unique. Then we construct a candidate set by pairing $M-K$ items with each size-$K$ bundle for a user, where the $M-K$ items are uniformly sampled from the whole items set. At last, we develop two datasets: 1) size-5 bundle along with $M$=100 as the size of the candidate set and 2) size-20 bundle with $M$=200 as the size of the candidate set. We name the above two datasets as Youshu (K=5, M=100) and Youshu (K=20, M=200).
    \item \textbf{Netease}. The dataset is crawled from Netease \citep{cao2017embedding}, a music streaming platform in which playlists containing various songs are seen as bundles. Due to a larger data size compared to the Youshu dataset, we first perform a sampling to obtain a relatively smaller dataset. Here, each bundle consists of at least 10 items and each item appears in at least 15 bundles. Further, each user in the dataset should be interacted with at least 15 bundles and 15 items. And following the same processing procedure as Youshu, we correspondingly construct a dataset called Netease (K=5, M=100).
\end{list}

\begin{table}[!htbp]
\vspace{-0.5cm}
\setlength{\abovecaptionskip}{0.5em}
\caption{The detailed information for datasets.}
\label{tab:dataset}
\renewcommand\arraystretch{1.1}
\begin{tabular}{lll}
\hline
                              & \textbf{Youshu}            & \textbf{Netease}          \\ \hline
\# Users                      & 8,006             & 2,532            \\
\# Bundles                    & 4,771             & 5,586            \\
\# Items                      & 32,770            & 51,298           \\
\# User-item Interactions     & 138,515           & 160,318          \\
\# User-bundle Interactions   & 51,377            & 75,536           \\
\# Bundle-item Affiliations   & 176,667           & 317,608          \\ \hline
\multicolumn{3}{l}{\textbf{Datasets for bundle generation}}                 \\ \hline
Youshu (K=5,N=100)             & \multicolumn{2}{l}{51,377 instances} \\
Youshu (K=20,N=200)            & \multicolumn{2}{l}{45,115 instances} \\
Netease (K=5,N=100)            & \multicolumn{2}{l}{75,536 instances} \\ \hline
\end{tabular}
\vspace{-0.5cm}
\end{table}

\subsection{Experimental Settings}
\emph{Evaluation protocol}. We perform an 80/20 split to construct the training and testing set for all datasets in line with \citep{gong2019exact,deng2021build}. To evaluate the performance of proposed model, three bundle-specific metrics are utilized \citep{deng2021build}:
\begin{equation}
    Precision@K=\frac{1}{|\boldsymbol{Y}|}\sum_{i}^{|\boldsymbol{Y}|}I(v_{\tilde{B}_{i}}^{(0)}=v_{Y_{i}}^{(0)})
\end{equation}
$Precision@K$ demonstrates whether the predicted next-item is the same as the one in the ground-truth bundle. Here, $Y=\{Y_i\},\widetilde{B}=\{\widetilde{B}_i\}$ are the set of ground-truth and generated bundles, respectively. $v_{*}^{(0)}$ denotes the first-place item in the bundle. $|\cdot|$ returns the size of the set, with $|Y|=|\widetilde B|$. $I(\cdot)$ is the indicator function and $K$ indicates the size of bundle.
\begin{equation}
    Precision^+@K=\frac{1}{|Y|}\sum_{i=1}^{|Y|}I(v_{\tilde{B}_i}^*\in Y_i)
\end{equation}
$Precision^+@K$ measures where there is an overlap between the generated bundle and the ground-truth bundle. $v_{\tilde{B}_i}^*$ stands for arbitrary item in the generated bundle $\tilde{B}_i$.
\begin{equation}
    Recall@K=\frac{1}{|Y|}\sum_{i=1}^{|Y|}\frac{\tilde{B}_i\cap Y_i}{K}
\end{equation}
$Recall@K$ indicates how many predicted items of $\tilde{B}_i$ are in the ground-truth bundle $Y_i$. Compared with $Precision^+@K$, $Recall@K$ is a more specific measure and gives a more prominent illustration on the generation quality.

\emph{Baselines}. We compare the proposed BundleNAT\footnote{The code will be released once accepted} with several competitive models in the experiments.
\begin{list}{\labelitemi}{\leftmargin=1em}
    \setlength{\topmargin}{0pt}
    \setlength{\itemsep}{0em}
    \setlength{\parskip}{0pt}
    \setlength{\parsep}{0pt}
    \item POP \citep{jeon2023accurate}: It chooses the top-\emph{k} popular items.
    \item BPR \citep{rendle2012bpr}: It is a well-known and effective pairwise ranking model, which is learned by optimizing the user-item pairwise ranking loss under the matrix factorization framework. The bundle here is generated based on top-K ranked user-item pairs from the candidate set.
    \item NCF \citep{he2017neural}: A competitive collaborative filtering method that combines the neural network architecture and traditional matrix factorization model to capture the non-linear interactions between users and items. We use the model to predict top-K most-likely to-interact-with items as the bundle for each user.
    \item UltraGCN \citep{mao2021ultragcn}: It is the ultra simplified GCN model for collaborative filtering, which further removes message passing on the basis of LightGCN and uses a constraint loss to approximate infinite-layer graph convolutions.
    \item SASRec \citep{kang2018self}: It is a sequential recommendation model based on self-attention mechanism which is capable of capturing long-range dependency so that high-quality prediction can be made when learning from user’s historical behaviors. To recover the size-K bundle, we employ the model to take turns to predict $K$ items.
    \item Exact-k \citep{gong2019exact}: This is the state-of-the-art model for building bundles which tries to find the maximal clique (bundle) in the graph composed by candidate items by employing an autoregressive-style encoder-decoder framework along with the demonstration mechanism from reinforcement learning.
    \item BYOB \citep{deng2021build}: It is a pure reinforcement-learning-based method recent proposed for bundle generation task, which tries to generate the personalized item set for users through the combination of proximal policy optimization and curriculum learning mechanism.   
\end{list}

BGGN \citep{chang2021bundle} and BGN \citep{bai2019personalized} mentioned in related works are excluded from the discussions due to the totally incompatible problem definition. Specifically, BGN is defined 
as a sequential recommendation task, the next bundle generation is merely based on historical user-bundle interactions; BGGN in fact treats the bundle generation as the graph completion task where items within a bundle are assumed to be sparsely related. We also do not include pre-built recommendation methods \citep{ma2022crosscbr,zhang2022suger} as baselines, which focus on finding the top-k bundles rather than generating a bundle from scratch.

\emph{Implementation details}. We choose Adaptive Moment Estimation (Adam) \citep{kingma2014adam} to optimize our BundleNAT. The dimension of the preference signal is set to be 64, and we use a 2-layer GNN to extract compatibility signal with an embedding size 128. Both encoding and decoding network is fixed to 2-block depth. Specifically, when utilizing a MF-BPR model for learning preference signal, a leave-one-out strategy \citep{he2016fast,sun2019bert4rec,petrov2022effective} is utilized to split user-item interactions for training. Grid search is adopted to find the best hyperparameter: the learning rate is searched within $\left\{10^{-4},10^{-3},10^{-2},10^{-1}\right\}$, the dropout ratio is tuned amongst $\left\{0.0,0.1,...,0.4\right\}$ and the coefficient of weight decay is in $\left\{10^{-5},10^{-4},...,10^{-1}\right\}$. All the experiments were conducted on a server with an AMD EPYC 7402 CPU and an NVIDIA GeForce RTX 3090 24G GPU. The server was running Ubuntu 22.04 with PyTorch \footnote{https://pytorch.org/} v1.9 and Python v3.6.

\subsection{Overall Performance Comparison (RQ1)}
We have tested the performance of the proposed BundleNAT against competitive baselines on three datasets, and the overall result is summarized in Table 2. The best and second-best results are emphasized by \textbf{bold} and \underline{underlined} fonts. We have observed the following findings:
\begin{table*}[]
\vspace{1.5em}
	\setlength{\abovecaptionskip}{0.25cm}
	\centering
	\caption{Performance comparison on three datasets.}
    \renewcommand\arraystretch{1.4}
    \resizebox{\linewidth}{!}{
    \begin{tabular}{cccccccccc}
        \hline
        \multicolumn{1}{c}{\textbf{Method}} & \multicolumn{3}{c}{\textbf{Youshu (K=5, M=100)}}                   & \multicolumn{3}{c}{\textbf{Youshu (K=20, M=200)}}                  & \multicolumn{3}{c}{\textbf{Netease (K=5, M=100)}}                  \\ \cline{3-3} \cline{6-6} \cline{9-9}
        \multicolumn{1}{l}{}                & Precision@5          & Precision+@5         & Recall@5             & Precision@20         & Precision+@20        & Recall@20            & Precision@5          & Precision+@5         & Recall@5             \\ \hline
        POP                                & 0.1000               & 0.8511               & 0.3949               & 0.0361               & 0.9917               & 0.5047               & 0.0493               & 0.6336               & 0.2046               \\
        BPR                                 & 0.4851               & 0.8313               & 0.3733               & 0.6425               & 0.9876               & 0.4745               & 0.2343               & 0.5691               & 0.1771               \\
        NCF                                 & 0.4943               & 0.8391               & 0.3770                & 0.6476               & 0.9862               & 0.4787               & \underline{0.2946}               & 0.6572               & 0.2297               \\
        UltraGCN                                 & 0.1039               & 0.8649               & 0.4025                & 0.0357              & 0.9907               & 0.4977               & 0.0639               & 0.6831               & \underline{0.2449}               \\
        SASRec                              & 0.4998               & 0.7884               & 0.2886               & 0.5875               & 0.9870                & 0.3578               & 0.2255               & 0.4920                & 0.1283               \\
        BYOB                                & 0.4700                 & 0.7900                 & 0.3042               & 0.6363               & 0.9797               & 0.1934               & 0.0595               & 0.4536               & 0.1070                \\
        Exact-k                             & \underline{0.5500}                 & \underline{0.8929}               & \underline{0.4307}               & \underline{0.7199}               & \underline{0.9918}               & \underline{0.5490}                & 0.2565               & \underline{0.6890}                & 0.2142               \\
        BundleNAT                                & \textbf{0.8091}               & \textbf{0.9582}               & \textbf{0.5843}               & \textbf{0.9779}               & \textbf{0.9996}               & \textbf{0.7566}               & \textbf{0.8551}               & \textbf{0.9451}               & \textbf{0.5937}               \\ \hline
        \multicolumn{1}{c}{Improved}       & \multicolumn{1}{c}{0.2591} & \multicolumn{1}{c}{0.0653} & \multicolumn{1}{c}{0.1536} & \multicolumn{1}{c}{0.2580} & \multicolumn{1}{c}{0.0078} & \multicolumn{1}{c}{0.2076} & \multicolumn{1}{c}{0.5605} & \multicolumn{1}{c}{0.2561} & \multicolumn{1}{c}{0.3488} \\ \hline
\end{tabular}
}
\end{table*}

Based on the empirical results, it can be seen that our method exceeds all the baselines by a large margin significantly. In detail, we can observe that absolute improvements in terms of Precision@5, Precision+@5, and Recall@5 are 25.91\%, 6.53\%, and 15.36\% against the second-best model, respectively, on Youshu datasets. As for the Netease dataset, the proposed BundleNAT still acquires a stably great performance, while the baselines apparently have difficulty in recovering the best bundle, with the absolute performance gains are 56.05\%/25.61\%/34.88\% on Precision@5/Precision+@5/Recall@5. Notably, existing baselines can gain a close performance regarding the Precision+@K metric, however, the poor performance on Recall@K metrics demonstrates the inferior ability to generate the exact bundle. The superiority of BundleNAT could be attributed to several factors: (1) BundleNAT employs the one-shot decoding manner which is fully aware of the order-invariant property; (2) BundleNAT effectively encodes the global dependency by incorporating the compatibility and preference signal.

It is surprising to find that POP gains a good performance on Precision+@k and Recall@k metric, which reveals that popular items are more likely to be chosen when forming a bundle. Our method outperforms the POP in every case, demonstrating that BundleNAT is able to generate bundles consisting of unpopular items as well as popular items.

BPR and NCF obtain close performance when compared with other specially designed models. The unexpected results suggest that the preference signal is informative for selecting the right item to composite the bundle, however, the significant performance gap indicates preference is not enough for high-quality generation.

Though UltraGCN shows remarkable performance in retrieving the desired bundle, e.g., obtaining the second-best performance on Recall@5 for the Netease dataset, its poor performance on the Precision metric indicates its ineffectiveness in finding the first appealing item in a bundle.

As shown in the Table 2, SASRec loses its superiority when applied to generate a bundle. A possible reason is that the inference of SASRec mainly focuses on modeling the sequential dependency between historical behaviors and the next interaction, while in bundle generation scenario we need to predict the next $K$ interactions. SASRec is not capable of capturing the dependency among the $K$ interactions, therefore resulting in poor performance in size-$K$ bundle generation. 

As far as we can see, Exact-k is the strongest competitor and obtains consistent good performance among all three datasets. When choosing an item, Exact-k takes the relations attended to preceding items into account via an attention-based module, consequently, the quality of the bundle is ensured. Meanwhile, BYOB generally falls behind the Exact-K on all the evaluation metrics, sometimes worse than other baselines. The performance degradation could be attributed to several reasons: 1) item compatibility information from the dataset is also leveraged to generate the bundle, however, the learning module based on word2vec is insufficient to capture a global dependency pattern due to fixed context size, consequently, the reinforcement learning framework is guided by a flawed signal; 2) the characteristics of real-world datasets are different from the synthetic ones for original BYOB evaluation, the specific-designed training order in curriculum learning could possibly fail in real-world bundle generation scenario. 

\begin{table*}[]\normalsize
\setlength{\abovecaptionskip}{0.25cm}
	\vspace{1em}
	\setlength{\abovecaptionskip}{0.25cm}
	\centering
	\caption{Performance comparison on Netease (K=5, M=100) regarding major designs.}
    \renewcommand\arraystretch{1.4}
    \begin{tabular}{lcccccc}
        \hline
        \textbf{Method}            & \textbf{Precision@5}                          &                                  & \textbf{Precision+@5}                          &                                  & \textbf{Recall@5}                               &         \\ \hline
        o/w   random      & 0.4610                                 & -46.09\%                          & 0.6165                                 & -34.77\%                          & 0.2415                                 & -59.32\% \\
        o/w   preference  & 0.4908                                 & -42.6\%                           & 0.6431                                 & -31.95\%                          & 0.2699                                 & -54.54\% \\
        o/w   compatibility & 0.8342                                 & -2.44\%                           & 0.9291                                 & -1.69\%                           & 0.5359                                 & -9.74\%  \\
        w/o   copy        & 0.7221                                 & -15.55\%                          & 0.8123                                 & -14.05\%                          & 0.3257                                 & -45.14\% \\
        r/w   max-pooling  & 0.8387                                 & -1.92\%                           & 0.9416                                 & -0.37\%                           & 0.5709                                 & -3.99\%  \\
        BundleNAT         & {\textbf{0.8551}} & {\textbf{}} & {\textbf{0.9451}} & {\textbf{}} & {\textbf{0.5937}} &         \\ \hline
\end{tabular}
\end{table*}

\subsection{Ablation Study (RQ2)}
We further conduct experiments on Netease (K=5, M=100) dataset to investigate the utility of various components in  BundleNAT. To better demonstrate their effectiveness, we develop several ablated models for comparison. Table 3 shows the performance of the default BundleNAT and ablated models.

\emph{Only with random feature (o/w random for short)}. The o/w random model refers to removing the preference and compatibility signal in the encoder, and only randomly initialized feature is used for generation. As shown in the Table 3, the o/w random model is generally the worst variant, which is not surprising since no informative characteristics are used for decoding.

\emph{Only with preference signal (o/w preference for short)}. The variant keeps preference signal and removes compatibility signal for encoding, thus the items within the bundle are selected based on users’ preference. It can be observed from the Table 3 that the utilization of preference signal can indeed find a better combination of items compared with o/w random variant, but still falls largely behind the default design, indicating that focus merely on preference is insufficient for bundle generation.

\emph{Only with compatibility signal (o/w compatibility for short)}. We combine the compatibility signal with the random initialized preference signal to formulate the encoder input. As we can see from the experimental result, the leverage of compatibility signal largely enhances the generation performance compared with o/w preference variant, demonstrating the necessity of properly encoding the item compatibility towards accurate bundle composition.

\emph{Without copy mechanism (w/o copy for short)}. Here, we remove the copy mechanism designed for decoder and the input of decoder is replaced by a randomly initialized trigger embedding \citep{wei2022towards}. Compared with the complete BundleNAT design, w/o copy variant experiences 15.55\%, 14.05\% and 45.14\% performance drop on Precision@5, Precision+@5 and Recall@5, respectively. It validates that a relatively high-quality start for decoding process is empirically beneficial, while from-scratch decoding has difficulty in recovering desired bundle.

\emph{Replace with max-pooling (r/w max-pooling for short)}. We replace the pooling function embedded in the copy mechanism with max-pooling method to obtain the decoder input. We can conclude from the result that r/w max-pooling is a good choice but still significantly inferior to the original design with regard to Precision@5 and Recall@5. Compared with the mean-pooling strategy, max-pooling tends to capture the most prominent local dependency which might cause global information loss. 

\subsection{Empirical Analysis on Time Efficiency (RQ3)}

In this section, we compare the proposed BundleNAT with the existing bundle-specific methods Exact-k and BYOB in terms of training time efficiency and inference latency. Figure 3 and Table 4 show the total training time and inference latency comparison, respectively, among the three models denoted by the multiplier.

In detail, we set the time cost and inference latency of BundleNAT as the base, Youshu-5, Youshu-20, and Netease-5 are the abbreviation for the three datasets. 

In Figure 3, we can observe that, when the target bundle size is 5 (Youshu-5, Netease-5), the second-best Exact-k is on average 7.99 times slower than our method. As the target bundle size grows, i.e., increases to 20, the time cost of Exact-k experiences a large increase due to the nonlinear expansion of search space, which becomes 32.273 times slower than BundleNAT. In addition, we can also find that, BundleNAT shows higher time efficiency than BYOB with average 2.03 times faster in overall runtime, although BYOB adopts a parallel speed-up computation framework \citep{moritz2018ray}  to speed up the RL-based inference process. 

\begin{figure*}[!h]
  \vspace{-0.25cm}
  \centering
  \includegraphics[scale=0.5]{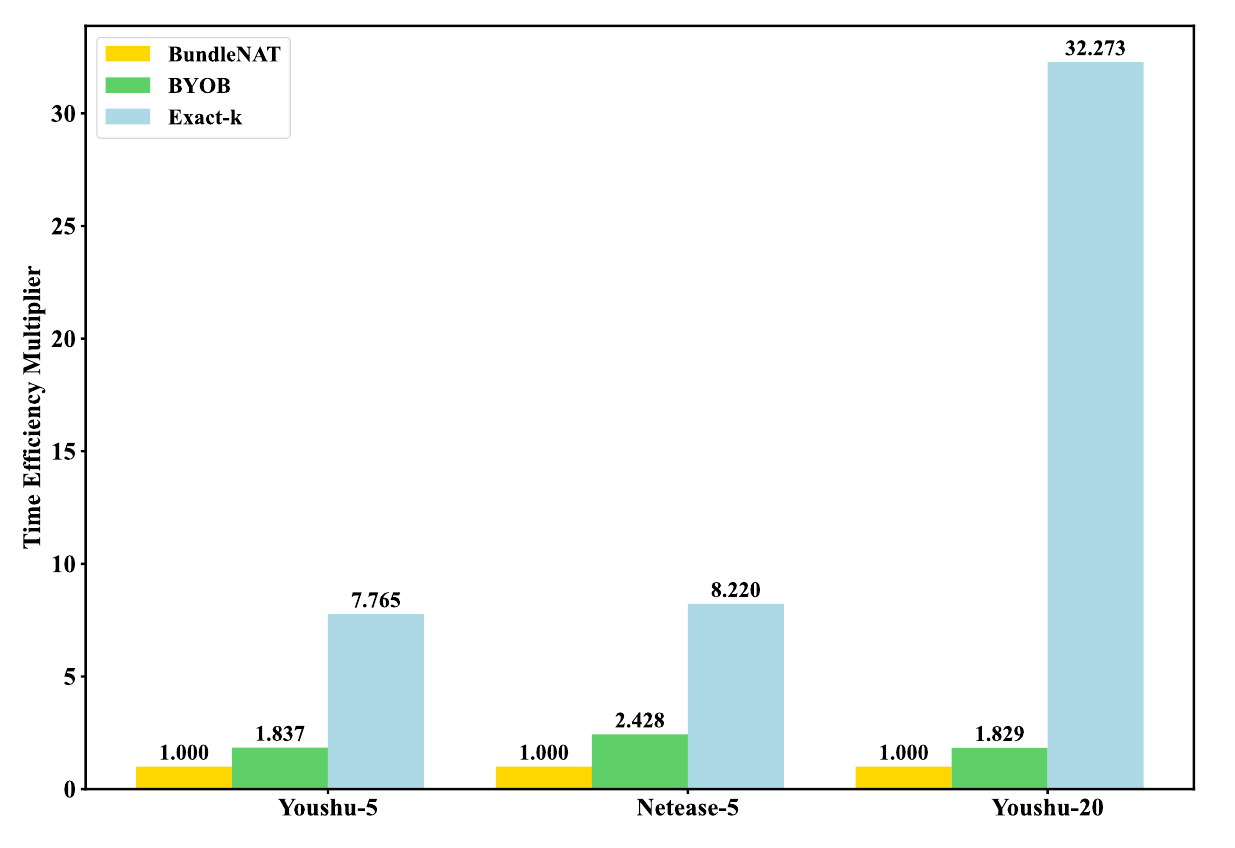}
  \caption{Overall training time comparison on all the three datasets. We compare the proposed BundleNAT with bundle-specific methods, i.e., Exact-k and BYOB.}
\end{figure*}

As we can see from Table 4, the BundleNAT fully exploits the inference advantage from non-autoregressive modeling. The time spent generating a bundle is significantly faster than Exact-k and BYOB. The BYOB has the worst inference efficiency among the models, and Exact-k has experienced a sharp increase as the bundle size gets larger.

The empirical results demonstrate the superior generation efficiency of the proposed non-autoregressive mechanism.

\begin{table*}[!t]
\vspace{1.5em}
	\setlength{\abovecaptionskip}{0.25cm}
	\centering
	\caption{Inference latency (per bundle) comparison on three datasets.}
\begin{tabular}{cccc}
\hline
\textbf{Method}    & \makecell{\textbf{Youshu}\\\textbf{(K=5, M=100)}}  & \makecell{\textbf{Youshu}\\\textbf{(K=20, M=200)}}  & \makecell{\textbf{Netease}\\\textbf{(K=5, M=100)}} \\ \hline
Exact-k   & 18.60x  & 126.47x & 16.47x  \\
BYOB      & 108.41x & 113.40x & 71.29x  \\
BundleNAT & 1.0x    & 1.0x    & 1.0x    \\ \hline
\end{tabular}
\end{table*}

\subsection{Parameters Analysis (RQ4)}
\begin{figure*}[!h]
  \centering
  \includegraphics[width=\linewidth]{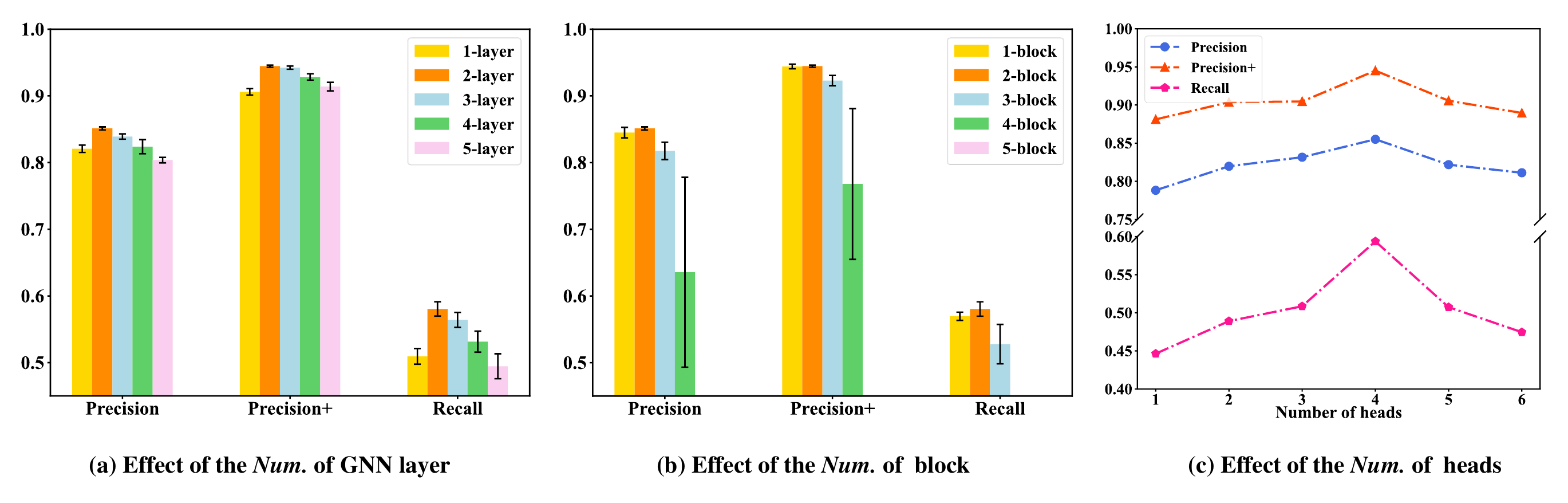}
  \caption{Influence of hyper-parameters on Netease (K=5, N=100) dataset.}
  \vspace{-0.25cm}
\end{figure*}

In this subsection, we perform sensitivity analysis of some important hyper-parameters in BundleNAT, including the number of layers of GNN to extract compatibility signal, the depth of encoding and decoding network, and the number of heads in the self-attention layer. We run 5 trials for each parameter during the analysis and report the averaged results in Figure 4.

\emph{The number of layers}. The number of GNN layers is searched in $\left\{1,2,3,4,5\right\}$. The corresponding results are presented in Figure 4(a). We can observe a 2-layer GNN is able to well retrieve higher-order signals and obtain a robust compatibility pattern with a relatively smaller standard deviation. As the number of GNN layers increases, the performance experiences continuous drop resulted from over-smoothing issue. It also can be seen that a 1-layer GNN is insufficient to capture the compatibility signal with only one-time information propagation.

\emph{The depth of encoding/decoding network}. The depth refers to number of encoding/decoding units in the Transformer architecture, which is searched from $\left\{1,2,3,4,5\right\}$. Figure 4(b) shows the empirical results. The BundleNAT obtains consistent performance gains as the depth increases, however, when the depth reaches a relatively larger number, i.e., greater than 2, the model starts to fall behind demonstrating that increased model complexity has no positive impact on model performance. When the number of encoding/decoding units is 2, it shows the greatest generation performance with the smallest deviation which illustrates the robustness of encoder-decoder architecture with depth 2.

\emph{The number of heads}. Here, we investigate the impact of the number of heads in self-attention module and tune among $\left\{1,2,3,4,5,6\right\}$. As we can see from Figure 4(c), single-head is not a preferred choice and results in poor performance without identifying various dependency patterns. Generally, the performance increases as the number of heads enlarges, while too many heads might begin to bring noisy information as the performance starts to fall when exceeding 4-head attention.

\section{Conclusion}
\subsection{Research Implications}
In this paper, we focus on the personalized bundle generation problem which aims to find the optimal bundle for users over a set of candidate items. Different from the previous study, we highlight the order-invariant property of the bundle and suggest following a sequential order is not suitable for generating the bundle resulting from inductive bias. We take the first step to formulate the bundle generation task via the non-autoregressive manner, and identify the corresponding challenges. To tackle this specific problem, we propose a novel encoder-decoder framework named BundleNAT. Specifically, we first design a self-attention based network to encode both preference signal and compatibility signal. Then we propose a non-autoregressive decoder to predict the targeted bundle in one-shot. We further propose a copy mechanism that facilitates the encoded pattern as the initial state of the decoder to ensure the generation towards the optimal solution. 

Extensive experiments on three real-world datasets demonstrate the effectiveness of the proposed BundleNAT, as BundleNAT significantly outperforms the existing state-of-the-art methods by a large margin. After time efficiency comparison, it can be seen that BundleNAT also shows significant advantages in generation efficiency. 

BundleNAT is closely related to the product bundling strategy in modern marketing and has many real-world applications. For example, on e-commerce platforms, it can select a set of appealing and compatible items from numerous candidate items for users efficiently and accurately based on historical purchases. It can also be effective on content platforms like music-streaming platform, it can deliver a satisfying playlist for listeners based on their likes. Besides, BundleNAT can perfectly fit with the real-time services in real-world practices due to the speed 
advantage brought by the non-autoregressive decoding.

\subsection{Future Work}
In future work, since there are various kinds of relationships among items, we will investigate how to incorporate heterogeneous information into dependency learning so that BundleNAT can extract a more complicated dependency pattern. Besides, the BundleNAT is currently designed for static scenario, i.e., the user's preference and item compatibility do not change/update over time, we plan to study the method for dynamic bundle generation.


\section*{Declaration of competing interest}
The authors declare that they have no known competing financial interests or personal relationships that could have appeared
to influence the work reported in this paper.

\section*{Acknowledgments}
This work was supported by the National Natural Science
Foundation of China (72025405, 72088101, 72001211), the National
Social Science Foundation of China (22ZDA102), the Hunan
Science and Technology Plan Project (2020TP1013, 2020JJ4673,
2022JJ20047, 2023JJ40685), the Shenzhen Basic Research Project
for Development of Science and Technology (JCYJ2020010914121
8676, 202008291726500001), and the Innovation Team Project of
Colleges in Guangdong Province (2020KCXTD040).

%\bibliographystyle{cas-model2-names}
\bibliographystyle{apacite}
\bibliography{cas-refs}

\begin{thebibliography}{}
\bibitem [\protect \citeauthoryear {%
Luo%
\ \protect \BOthers {.}}{%
Luo%
\ \protect \BOthers {.}}{%
{\protect \APACyear {2022}}%
}]{%
luo2022your}
\APACinsertmetastar {%
luo2022your}%
\begin{APACrefauthors}%
Luo, S.%
, Li, S.%
, Zheng, S.%
, Liu, T\BHBI Y.%
, Wang, L.%
\BCBL {}\ \BBA {} He, D.%
\end{APACrefauthors}%
\unskip\
\newblock
\APACrefYearMonthDay{2022}{}{}.
\newblock
{\BBOQ}\APACrefatitle {Your transformer may not be as powerful as you expect}
  {Your transformer may not be as powerful as you expect}.{\BBCQ}
\newblock
\APACjournalVolNumPages{arXiv preprint arXiv:2205.13401}{}{}{}.
\PrintBackRefs{\CurrentBib}
\end{thebibliography}


%\linenumbers
\end{document}